\documentclass{article} 
\usepackage[preprint]{colm2026_conference}
\usepackage{microtype}
\usepackage{hyperref}
\usepackage{url}
\usepackage{booktabs}

\usepackage{lineno}

\definecolor{darkblue}{rgb}{0, 0, 0.5}
\hypersetup{colorlinks=true, citecolor=darkblue, linkcolor=darkblue, urlcolor=darkblue}

\usepackage{hyperref}
\usepackage{url}
\usepackage{booktabs}
\usepackage{amsmath,amssymb,amsfonts}
\usepackage{graphicx}
\usepackage{xcolor}
\usepackage{xspace}
\usepackage{multirow}
\usepackage{cleveref}
\usepackage{algorithm}
\usepackage{algorithmic}
\usepackage{lineno}
\usepackage{pifont}
\usepackage[most]{tcolorbox}
\usepackage{tikz}
\newtcolorbox{observationbox}{
  colback=blue!4,
  colframe=blue!40!black,
  boxrule=0.5pt,
  arc=2pt,
  left=6pt, right=6pt, top=6pt, bottom=6pt,
  fonttitle=\bfseries,
  title=Key Observation
}

\definecolor{darkblue}{rgb}{0, 0, 0.5}
\hypersetup{colorlinks=true, citecolor=darkblue, linkcolor=darkblue, urlcolor=darkblue}

\newcommand{\method}{SODA\xspace}

\newcommand{\eg}{e.g.\xspace}

\title{SODA: Semi On-Policy Black-Box Distillation for Large Language Models}

\author{\textbf{Xiwen Chen}$^{1}$\footnotemark[1]\thanks{Equal contribution},
 \textbf{Jingjing Wang}$^{1}$\footnotemark[1],
\textbf{Wenhui Zhu}$^{2}$\footnotemark[1],
 \textbf{Peijie Qiu}$^{3}$, 
 \textbf{Xuanzhao Dong}$^{4}$, \\
 \textbf{Yueyue Deng}$^{5}$, 
 \textbf{Hejian Sang}$^{2}$, 
  \textbf{Zhipeng Wang}$^{2}$\footnotemark[1] , 
  \textbf{Alborz Geramifard}$^{2}$,
   \textbf{Feng Luo}$^{1}$
  \\
  \\
  $^1$Clemson University, $^2$LinkedIn, $^3$Washington University in St. Louis, \\
   $^4$Arizona State University,  $^5$Columbia University \\
  \texttt{xiwenchen.cs@outlook.com, jingjiw@g.clemson.edu, wenhzhu@linkedin.com} \\
}

\begin{document}

\ifcolmsubmission
\linenumbers
\fi

\maketitle

\begin{figure}[!htb]
\centering
\includegraphics[width=1.0\linewidth]{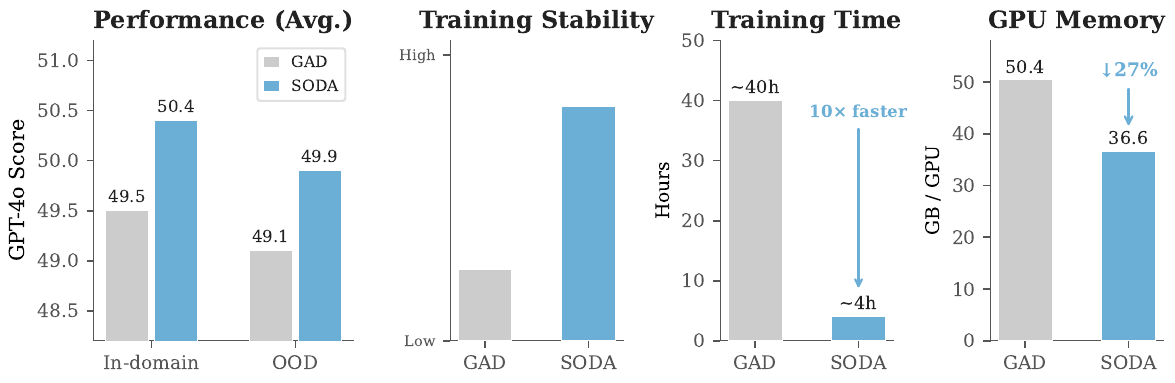}
\vspace{-0.2in}
\caption{\method achieves competitive or better distillation quality than GAD: 10$\times$ faster and 27\% more memory-efficient, while being substantially easier and more stable to train.
From left to right: GPT-4o Score averaged over four student models (higher is better; 50 denotes GPT-4o parity); training stability; wall-clock training time; and peak GPU memory.
}
\label{fig:teaser}
\end{figure}

\begin{abstract}
Black-box knowledge distillation for large language models presents a strict trade-off. Simple off-policy methods (e.g., sequence-level knowledge distillation) struggle to correct a student's inherent errors. Fully on-policy methods (e.g., Generative Adversarial Distillation) solve this via adversarial training but introduce well-known instability and crippling computational overhead. To address this dilemma, we propose \method (\textbf{S}emi \textbf{O}n-policy \textbf{D}istillation with \textbf{A}lignment), a highly efficient alternative that leverages the inherent capability gap between frontier teacher models and much smaller students. Since a compact student model's natural, zero-shot responses are almost strictly inferior to a powerful teacher's outputs, we construct a highly effective contrastive signal simply by pairing the teacher's responses with a raw, one-time static snapshot of the student's rollouts. By seamlessly translating this gap into a concise preference optimization pipeline, we expose the student to its own static inferior modes. This achieves high-quality distribution alignment while completely eliminating the need for complex filtering, costly dynamic rollouts, or fragile adversarial training. Extensive evaluations across four compact Qwen2.5 and Llama-3 models validate this semi on-policy paradigm. \method matches or outperforms state-of-the-art methods on 15 out of 16 benchmark settings. More importantly, it achieves this comparable distillation quality while training \textbf{10$\times$} faster, consuming \textbf{27\%} less peak GPU memory, and entirely bypassing adversarial instability.
\end{abstract}

\section{Introduction}
\label{sec:intro}
Knowledge distillation~\citep{kd} of frontier large language models (LLMs;~\citealp{gpt5}) has become the primary paradigm for creating capable, efficient \emph{small models} that are practical to deploy~\citep{qwen2.5,llama3}. 
When the teacher is a proprietary frontier model (\eg, GPT-5 or Claude), its internal parameters, hidden states, and continuous logit distributions are strictly inaccessible. Practitioners can only query the model through an API to obtain discrete text generations. 
This highly constrained setting, known as \emph{black-box distillation}, fundamentally diverges from \emph{white-box distillation} (where full access to the teacher's probability space is assumed). Consequently, powerful white-box alignment techniques that rely on minimizing KL divergence or matching internal representations are entirely inapplicable. 
The \textit{de facto} standard in this discrete regime is sequence-level knowledge distillation (SeqKD;~\citealp{skd}), which fine-tunes the small student model on these teacher outputs via supervised learning~\citep{alpaca,vicuna,ITGPT4,lima}.
While simple and highly scalable, SeqKD suffers from a fundamental flaw: it is purely \emph{off-policy}. The student passively imitates teacher demonstrations without any exposure to its own generative distribution, leaving it unaware of its innate inferior tendencies. This mismatch severely limits out-of-distribution generalization~\citep{sftmemrlgen}.
Recent work highlights the importance of \emph{on-policy} learning~\citep{minillm,googlepolicy}. Extending this idea to the black-box regime, Generative Adversarial Distillation (GAD;~\citealp{ye2025black}) brings fully on-policy learning via a minimax game~\citep{gan,seqgan}. 
However, GAD introduces immense computational and architectural overhead: it requires maintaining an additional discriminator network of comparable size, performing alternative generator-discriminator updates, and balancing fragile adversarial training dynamics. However, for researchers and practitioners aiming to efficiently train \emph{small models}, the prohibitive resource requirements of fully on-policy adversarial distillation largely defeat the purpose of efficiency. 

This dilemma raises a fundamental question: \emph{is fully on-policy, continuous feedback strictly necessary, or can we retain the benefits of student-aware error correction without the overhead of adversarial training?}

In this work, we demonstrate that adversarial training is unnecessary to achieve effective distribution alignment (see~\Cref{fig:teaser}). Instead, we propose a much simpler and highly efficient alternative motivated by a key observation that, given the inherent capability gap between a frontier teacher and a small base model, the student's natural, zero-shot responses are almost strictly inferior to the teacher's responses. Leveraging this natural contrast, we introduce \method (\textbf{S}emi \textbf{O}n-policy \textbf{D}istillation with \textbf{A}lignment). 
\method seamlessly translates this static capability gap into an elegant preference optimization pipeline. Starting with a brief warmup on the teacher data to stabilize the initial policy, \method directly applies Direct Preference Optimization (DPO;~\citealp{rafailov2024direct}) using the teacher's responses as preferred and the base small model's own natural responses as dispreferred. This concise formulation yields a powerful dual learning signal: teacher imitation (learning the target behavior) and mode pruning (suppressing the small model's innate errors).

We characterize this method as \emph{semi on-policy}: unlike SeqKD, \method heavily incorporates information about the student's own distribution; unlike GAD, it draws this signal from a one-time static snapshot, bypassing the need for expensive online sampling. By decoupling the contrastive signal from the training loop, \method eliminates the discriminator and adversarial RL entirely. This leads to a $10\times$ speedup over GAD, effectively demonstrating that a targeted, static snapshot of the student's inferior behaviors is sufficient for high-quality distillation, eliminating the need for continuous online tracking.
We validate \method using GPT-5-Chat~\citep{gpt5} as the teacher across four open-source small models from the Qwen2.5~\citep{qwen2.5} and Llama-3~\citep{llama3} families (3B--14B parameters) on the LMSYS-Chat dataset~\citep{lmsys}. 

Our contributions can be summarized as follows: (i) We introduce the concept of \emph{semi on-policy} distillation, demonstrating that a static snapshot of a small student model's prior distribution provides an extremely effective, targeted contrastive signal for black-box alignment. (ii) We propose \method, an elegant and lightweight distillation pipeline that corrects student-specific errors without the need for additional models or continuous adversarial sampling. (iii) Extensive evaluations show that \method achieves comparable or slightly improved distillation quality relative to the state-of-the-art GAD (matching or exceeding it on 15 out of 16 model--dataset combinations). Most importantly, \method delivers this strict performance parity while being 10$\times$ faster and consuming 27\% less peak GPU memory (\Cref{fig:teaser}).

\section{Related Work}
\label{sec:related}

\noindent\textbf{White-Box On-Policy Distillation.} 
Recent literature demonstrates that on-policy learning is crucial for correcting a student's inherent generation errors. However, these foundational on-policy methods operate strictly in the \textit{white-box} regime. For example, MiniLLM \citep{gu2024minillm} minimizes the reverse Kullback-Leibler divergence, while generalized frameworks \citep{agarwal2024policy} and mechanistic recipes \citep{li2026rethinking} compute explicit token-level feedback on student-generated trajectories. Crucially, all these approaches require full transparent access to the proprietary teacher's internal logits and continuous probability distributions to score the student's on-policy outputs. Consequently, they are fundamentally inapplicable to the \textit{black-box} setting, where practitioners can only query frontier APIs to obtain discrete text.

\noindent\textbf{Black-Box Distillation.} 
In this highly constrained regime, the standard approach is sequence-level supervised fine-tuning (SeqKD; \citealp{skd}), used by models like Alpaca, Vicuna, and LIMA \citep{alpaca,vicuna,ITGPT4,lima}. While simple, SeqKD is purely off-policy: the student only imitates the teacher and ignores its own innate errors \citep{f-div-kd}. To bring on-policy learning to the black-box setting, Generative Adversarial Distillation (GAD; \citealp{ye2025black}) uses a discriminator \citep{gan,seqgan} to align student and teacher outputs from a distribution perspective. However, GAD suffers from severe training instability and prohibitive memory costs. Our method, \method, sits between these two. It captures the essential on-policy contrastive signal using a one-time static snapshot of the student outputs, entirely eliminating the immense overhead of adversarial training.

\noindent\textbf{Preference Optimization as Distillation.} 
DPO \citep{rafailov2024direct} is a common way to align LLMs. While usually applied for general goals like safety or helpfulness, recent works have increasingly explored preference optimization for model distillation. For instance, Zephyr \citep{tunstall2023zephyr} and DPKD \citep{li2024direct} employ DPO primarily to transfer alignment and instruction-following capabilities from teachers to students. Closest to our data setup is PLaD \citep{zhang2024plad}, which constructs pseudo-preference pairs from teacher and student outputs to calibrate sequence likelihoods via a ranking loss. However, these existing pipelines typically rely on external judge models for filtering, iterative sampling, or complex modified objectives. In contrast, \method is explicitly designed as an ultra-efficient substitute for adversarial black-box distillation (e.g., GAD). We demonstrate a minimalist finding: the raw, unfiltered static snapshot of the base student's ($q_0$) zero-shot responses provides a sufficiently precise contrastive signal. By pairing these with teacher responses, \method leverages DPO specifically as a mode-pruning mechanism. This naturally corrects the student's characteristic inferior behaviors without introducing additional models, filtering heuristics, or continuous online sampling. 

\section{Method}
\label{sec:method}

We consider the problem of \emph{black-box} knowledge distillation for large language models (LLMs).
A student model $q_\theta(y \mid x)$ is trained to approximate a proprietary teacher $p(y \mid x)$, given only the teacher's text responses; no logits, gradients, or internal representations are accessible.
The distillation dataset $\mathcal{T} = \{(x_i, y_i^t)\}_{i=1}^{N}$ consists of prompts $x_i$ paired with teacher-generated responses $y_i^t \sim p(\cdot \mid x_i)$.
This black-box constraint is the practical reality when distilling from proprietary models such as GPT-5 or Claude, where only API access to generated text is available.

\subsection{Preliminaries}
\label{sec:prelim}

\noindent\textbf{Sequence-level knowledge distillation.}
The dominant approach to black-box distillation is \emph{sequence-level knowledge distillation} (SeqKD; \citealp{skd}), which performs supervised fine-tuning (SFT) on teacher-generated text:
\begin{equation}
    \mathcal{L}_{\text{SFT}}(\theta) = -\mathbb{E}_{(x, y^t) \sim \mathcal{T}} \left[ \log q_\theta(y^t \mid x) \right],
    \label{eq:sft}
\end{equation}
where the loss is computed only on assistant response tokens, with all prompt tokens masked.
Starting from a pre-trained student $q_0$ (\eg Qwen2.5-7B-Instruct), minimizing $\mathcal{L}_{\text{SFT}}$ yields the SFT model $q_{\text{SFT}}$.
SeqKD is simple, stable, and widely adopted \citep{alpaca,vicuna,ITGPT4,lima} as the de facto black-box distillation baseline.

\noindent\textbf{Generative Adversarial Distillation.}
GAD \citep{ye2025black} is a recent method that brings \emph{on-policy} learning to the black-box setting through adversarial training.
It frames the student as a generator $G$ and introduces a discriminator $D$ that assigns a sequence-level scalar score $D(y)$ to a response $y$.
The training objective is a minimax game with value function:
\begin{equation}
    \max_G \min_D \ \mathcal{V}(G, D) = \mathbb{E}_{(x, y^t) \sim \mathcal{T}} \left[ -\log \sigma\!\left( D(y^t) - D(G(x)) \right) \right],
    \label{eq:gad}
\end{equation}
where $\sigma$ is the sigmoid function and the Bradley-Terry model \citep{bradley} captures pairwise preferences.
The generator is optimized via policy gradient to maximize $D(G(x))$, while the discriminator is trained to score teacher responses higher than student responses.
GAD requires a warmup stage (one epoch of SFT for the generator and Bradley-Terry training for the discriminator) before adversarial training begins.

\subsection{Limitations of Existing Approaches}
\label{sec:limitations}

\noindent\textbf{Why is SeqKD insufficient?}
SeqKD is a purely \emph{off-policy} method: the student learns exclusively from the teacher's demonstrations, with no information about its own generation behavior.
This leads to two fundamental limitations.
First, the student receives only \emph{positive} signal: it learns what good responses look like, but never learns \emph{what to avoid}.
Standard SFT has no mechanism for incorporating negative examples; the student cannot contrast good teacher behavior against its own characteristic errors.
Second, SeqKD suffers from \emph{exposure bias} \citep{exposure_bias}: during training the student is conditioned on ground-truth teacher prefixes, but at inference time it must condition on its own (potentially flawed) generations.
This train-test mismatch compounds across long sequences, as errors in early tokens propagate to later ones.
These limitations are well-documented in the white-box setting, where on-policy methods, via reverse KLD \citep{minillm} or generalized divergences \citep{f-div-kd}, consistently outperform off-policy SeqKD.
The question is how to realize similar benefits in the black-box setting, where the teacher's probability space is entirely inaccessible.

\noindent\textbf{Why is fully on-policy distillation problematic?}
GAD (\cref{eq:gad}) addresses the above limitations by introducing a discriminator that provides on-policy feedback, but it inherits the well-known difficulties of adversarial training.
The minimax objective requires careful balancing between generator and discriminator updates: if the discriminator becomes too strong, the reward signal saturates and gradients vanish; if too weak, the feedback becomes uninformative.
Beyond stability, GAD incurs substantial computational overhead.
It maintains and trains a separate discriminator network (initialized from the student), and optimizes the generator with GRPO \citep{grpo}, which samples \emph{multiple} completions per prompt at every training step to estimate the baseline reward.
This means \emph{on-policy generation is not a single forward pass but a full rollout of $K$ responses per prompt per update, compounding the cost of sequence generation on top of the already doubled memory footprint from the discriminator.}
The warmup schedule, learning rate ratio between generator and discriminator, number of rollouts $K$, and other RL hyperparameters all require careful tuning, with failure modes that are difficult to diagnose.
These issues raise a natural question: \emph{is the full complexity of on-policy adversarial training necessary, or can a simpler approach capture most of its benefit?}

\subsection{\method: Semi On-Policy Distillation with Alignment}
\label{sec:soda}

The limitations above share a common root: SeqKD ignores the student's own distribution entirely, while GAD pays a heavy price to track it continuously.
\begin{observationbox}
The standard black-box distillation setup already contains a natural preference signal that requires no human annotation, reward modeling, or adversarial tracking.
Given the inherent capability gap between the models, the teacher's response provides an optimal target, while a single zero-shot forward pass through the base student naturally reveals the inferior outputs \emph{this particular student} would produce instead. 
The quality gap between the two forms a structured, student-specific contrastive signal, explicitly exposing the behaviors the student must learn to suppress.
\end{observationbox}

\noindent \method exploits this signal: by pairing teacher outputs against the student's own responses and optimizing with a preference objective, we turn the teacher--student gap into a targeted alignment curriculum, at essentially zero additional cost beyond the distillation data itself.

Concretely, we sample responses from the base student $q_0$ \emph{before any fine-tuning}:
\begin{equation}
    y_i^{s} \sim q_0(\cdot \mid x_i), \quad i = 1, \ldots, N,
    \label{eq:student_gen}
\end{equation}
and pair them against the teacher responses to form a preference dataset:
\begin{equation}
    \mathcal{D}_{\text{pref}} = \left\{ \left(x_i,\; y_i^{+} = y_i^{t},\; y_i^{-} = y_i^{s}\right) \right\}_{i=1}^{N}.
    \label{eq:pref_data}
\end{equation}
Each pair encodes a direct contrast between \emph{where the teacher is} and \emph{where the student currently stands}.
This is what makes \method \emph{semi on-policy}: the negative signal comes from the student's own distribution (on-policy), but is captured once and held fixed (off-policy in the temporal sense).

The core of \method is to distill the teacher's behavior into the student through preference optimization on $\mathcal{D}_{\text{pref}}$.
Prior to this, we warm up the student by briefly fine-tuning $q_0$ on teacher responses (\cref{eq:sft}) to obtain a reasonable initialization $q_{\text{w}}$.
Starting from $q_{\text{w}}$, we apply Direct Preference Optimization (DPO; {\citealp{rafailov2024direct}}) on $\mathcal{D}_{\text{pref}}$:
\begin{equation}
    \mathcal{L}_{\text{DPO}}(\theta) = -\mathbb{E}_{(x, y^+, y^-) \sim \mathcal{D}_{\text{pref}}} \left[ \log \sigma \!\left( \beta \log \frac{q_\theta(y^+ \mid x)}{q_{\text{w}}(y^+ \mid x)} - \beta \log \frac{q_\theta(y^- \mid x)}{q_{\text{w}}(y^- \mid x)} \right) \right],
    \label{eq:dpo}
\end{equation}
where $q_{\text{w}}$ serves as the reference policy and $\beta$ controls the KL regularization strength.
The warmup merely provides a stable starting point; the distillation itself happens through preference optimization on $\mathcal{D}_{\text{pref}}$, which teaches the student \emph{what to stop doing}---namely the characteristic behaviors of its own prior $q_0$ that diverge from the teacher.

\noindent\textbf{Why must the negatives come from the student itself?}
One might ask: why not construct $\mathcal{D}_{\text{pref}}$ using any source of low-quality responses, such as outputs from a weaker unrelated model, synthetically corrupted text, or even random samples?
The answer is that generic negatives encode what is bad \emph{in general}, but carry no information about what \emph{this particular student} gets wrong.
DPO's gradient pushes probability mass away from rejected responses and toward preferred ones.
When the rejected responses are drawn from $q_0$, this gradient is concentrated on the regions of output space that the student is \emph{actually likely to visit}, i.e., its own suboptimal distribution.
Generic negatives, by contrast, may occupy regions the student would never reach in practice, wasting optimization effort on irrelevant corrections.
In other words, student-sourced negatives ensure that every gradient step addresses a \emph{real} inferior behavior, not a hypothetical one.
Beyond effectiveness, using $q_0$ as the negative source is also the most practical choice: the base student is already available as the starting point for training, so generating its responses requires no additional model, no API cost, and no external data. It requires only a single batched inference pass over the training prompts.

\noindent\textbf{Practical considerations.}
Sampling from $q_0$ (\cref{eq:student_gen}) is the only additional step beyond the standard distillation data; it is embarrassingly parallel and runs offline via batched inference (we use vLLM; {\citealp{kwon2023vllm}}), adding negligible overhead relative to training.
Since the student responses are generated before training begins, preference dataset construction requires no interruption of the training pipeline.
No additional model is introduced: \method uses only the student architecture throughout, in contrast to GAD's separate discriminator.

\noindent\textbf{Relation to standard alignment pipelines.}
While \method adopts the optimization framework of DPO, its objective deviates from standard RLHF. 
Standard RLHF maximizes a latent reward based on human preferences, where rejected responses serve as generic boundaries. 
In contrast, we treat distillation as a distribution alignment problem. 
Traditional token or sequence-level distillation often struggles to bridge the distributional gap, particularly in reasoning tasks involving long generation trajectories. 
\method addresses this by reformulating alignment as a preference learning task. 
By contrasting teacher samples with those from the student prior ($q_0$), we isolate the specific regions where the student diverges. 
This method enables effective distribution alignment, achieving better reasoning performance than SFT while avoiding the instability of adversarial training.

\subsection{The Semi On-Policy Perspective}
\label{sec:spectrum}

We situate \method within a spectrum of black-box distillation methods, characterized by how much student-distribution information the distillation signal incorporates.
\textbf{Off-policy} methods (SeqKD) learn exclusively from teacher demonstrations $y^t \!\sim\! p$, with no information about the student's own behavior.
\textbf{Semi on-policy} methods (\method) additionally incorporate the student's prior distribution $q_0$ as a static contrastive signal, student-specific but fixed before training begins.
\textbf{Fully on-policy} methods (GAD) continuously sample from and evaluate the \emph{current} student $q_\theta$, co-evolving the feedback signal at every training step.

Moving along this spectrum increases the relevance of the feedback signal to the student's current policy, but also increases computational cost and training complexity (\cref{tab:compute}).
The central claim of this work is that \emph{most of the benefit of on-policy feedback can be captured by a one-time snapshot of the student's distribution}, without requiring continuous co-adaptation.
The intuition is that the base student $q_0$ and the training-time student $q_\theta$ share the same architecture, pretraining, and inductive biases.
Many of the systematic biases present in $q_0$ (verbose completions, hallucinated facts, stylistic tics) persist in attenuated form throughout training.
By penalizing $q_0$'s characteristic outputs through preference optimization, \method applies corrective pressure on precisely these persistent inferior behaviors, achieving targeted penalization without adversarial co-evolution.

An important corollary is that the semi on-policy signal is \emph{front-loaded}: it is most informative when $q_\theta$ is still close to $q_0$ (early in preference training), and its utility diminishes as $q_\theta$ diverges.

\noindent\textbf{Theoretical analysis of the dual learning signal.}
The effectiveness of base student negatives can be understood through the connection between DPO and inverse reinforcement learning (IRL) under the maximum entropy framework~\citep{maxent_irl}.
DPO implicitly recovers a reward function from preference data without explicit reward modeling~{\citep{rafailov2024direct}}:
\begin{equation}
    r^*(x, y) = \beta \log \frac{q_{\text{SODA}}(y \mid x)}{q_{\text{w}}(y \mid x)} + \beta \log Z(x),
    \label{eq:implicit_reward}
\end{equation}
where $q_{\text{SODA}}$ is the converged policy and $Z(x)$ is a per-prompt partition function.
In the \method setting, the preference data pairs teacher responses against the student's own prior outputs, so $r^*$ encodes \emph{what makes the teacher's behavior better than this particular student's}, a reward signal intrinsically calibrated to the student's inferior behaviors, analogous to IRL recovering a reward from expert--novice demonstration contrasts.
The converged policy equivalently solves the KL-regularized objective~{\citep{rafailov2024direct}}:
\begin{equation}
    q_{\text{SODA}} = \arg\max_{q_\theta}\; \mathbb{E}_{x}\!\left[\mathbb{E}_{y \sim q_\theta}\!\big[r^*(x, y)\big] - \beta\, \mathrm{KL}\!\big(q_\theta(\cdot \mid x) \,\big\|\, q_{\text{w}}(\cdot \mid x)\big)\right].
    \label{eq:rl_objective}
\end{equation}
The first term drives $q_\theta$ toward high-reward (teacher-like) outputs and away from low-reward ($q_0$-like) outputs; the second anchors $q_\theta$ near $q_{\text{w}}$, preventing catastrophic drift.
Together, they yield a policy that selectively suppresses the base student's characteristic inferior behaviors while reinforcing teacher-like behavior---the dual learning signal that constitutes \method's distillation mechanism.

\noindent\textbf{Gradient concentration on the student's support.}
The mechanism is visible in the DPO gradient for a single preference pair $(x, y^+, y^-)$:
\begin{equation}
    \nabla_\theta \mathcal{L}_{\text{DPO}} = -\beta\, \underbrace{\sigma\!\big(\!-\!\hat{r}_\theta\big)}_{\text{adaptive weight}} \Big[\underbrace{\nabla_\theta \log q_\theta(y^+ \!\mid\! x)}_{\text{imitate teacher}} \;-\; \underbrace{\nabla_\theta \log q_\theta(y^- \!\mid\! x)}_{\text{suppress rejected}}\Big],
    \label{eq:dpo_grad}
\end{equation}
where $\hat{r}_\theta \!=\! \beta \log \frac{q_\theta(y^+|x)}{q_{\text{ref}}(y^+|x)} - \beta \log \frac{q_\theta(y^-|x)}{q_{\text{ref}}(y^-|x)}$ is the implicit reward margin.
When $y^- \!\sim\! q_0$, the ``suppress rejected'' term $\nabla_\theta \log q_\theta(y^- \!\mid\! x)$ is the score function evaluated at sequences the student naturally produces.
Since $q_\theta$ (initialized from $q_{\text{w}}$, itself derived from $q_0$) assigns non-negligible probability to these outputs, the score function is well-conditioned and each gradient step meaningfully reshapes the student's distribution in the regions it actually visits.
The combined effect is a dual learning signal: the teacher-positive term performs \emph{teacher imitation} (learning what the teacher produces), while the $q_0$-negative term performs \emph{mode pruning} (suppressing the student's prior inferior behaviors); see \Cref{fig:dual_signal} for an illustration.
Pure imitation learning (SeqKD) achieves only the former; \method achieves both through preference optimization, which explains why the preference distillation phase yields consistent gains over imitation alone (\cref{sec:experiments}).

\begin{figure}[t]
\centering
\resizebox{0.8\textwidth}{!}{%
\begin{tikzpicture}[>=stealth, font=\small]

\begin{scope}
  \node[font=\small\bfseries, anchor=south] at (3.2, 3.5) {(a) Warmup};
  \draw[->, black!35, thick] (-0.2, 0) -- (6.6, 0);
  \draw[->, black!35, thick] (0, -0.1) -- (0, 3.3);
  \node[black!45, font=\scriptsize] at (3.3, -0.35) {response quality};

  \fill[blue!8]
    plot[smooth, domain=2.8:6.4, samples=50]
      (\x, {2.8*exp(-(\x-5)*(\x-5)/0.6)})
    -- (6.4, 0) -- (2.8, 0) -- cycle;
  \draw[blue!60!black, line width=1.3pt]
    plot[smooth, domain=1:6.5, samples=70]
      (\x, {2.8*exp(-(\x-5)*(\x-5)/0.6)});
  \node[blue!60!black, font=\footnotesize] at (5.55, 2.5) {$p$};

  \draw[red!60!black, line width=1.3pt]
    plot[smooth, domain=0:5.5, samples=70]
      (\x, {1.8*exp(-(\x-2)*(\x-2)/0.9)});
  \node[red!60!black, font=\footnotesize] at (1.15, 1.65) {$q_0$};

  \draw[orange!70!black, line width=1.3pt, dashed]
    plot[smooth, domain=0.2:6.5, samples=90]
      (\x, {1.9*exp(-(\x-3.8)*(\x-3.8)/0.65) + 0.5*exp(-(\x-2.1)*(\x-2.1)/0.55)});
  \node[orange!70!black, font=\footnotesize] at (4.55, 2.1) {$q_{\text{w}}$};

  \draw[->, red!45!black, thin] (3.2, 1.05) -- (2.45, 0.6);
  \node[red!45!black, font=\tiny, anchor=south] at (3.3, 1.05) {residual modes};

  \draw[->, blue!50!black, line width=1.2pt] (2.2, 2.9) -- (4.2, 2.9);
  \node[blue!50!black, font=\scriptsize, above=-1pt] at (3.2, 2.9) {teacher imitation};
\end{scope}

\begin{scope}[xshift=7.8cm]
  \node[font=\small\bfseries, anchor=south] at (3.2, 3.5) {(b) Preference Distillation};
  \draw[->, black!35, thick] (-0.2, 0) -- (6.6, 0);
  \draw[->, black!35, thick] (0, -0.1) -- (0, 3.3);
  \node[black!45, font=\scriptsize] at (3.3, -0.35) {response quality};

  \fill[blue!8]
    plot[smooth, domain=2.8:6.4, samples=50]
      (\x, {2.8*exp(-(\x-5)*(\x-5)/0.6)})
    -- (6.4, 0) -- (2.8, 0) -- cycle;
  \draw[blue!60!black, line width=1.3pt]
    plot[smooth, domain=1:6.5, samples=70]
      (\x, {2.8*exp(-(\x-5)*(\x-5)/0.6)});
  \node[blue!60!black, font=\footnotesize] at (5.55, 2.5) {$p$};

  \draw[orange!70!black, line width=1pt, dashed, opacity=0.35]
    plot[smooth, domain=0.2:6.5, samples=90]
      (\x, {1.9*exp(-(\x-3.8)*(\x-3.8)/0.65) + 0.5*exp(-(\x-2.1)*(\x-2.1)/0.55)});
  \node[orange!70!black, font=\footnotesize, opacity=0.45] at (3.3, 1.5) {$q_{\text{w}}$};

  \fill[red!15, opacity=0.55]
    plot[smooth, domain=0.5:3.8, samples=40]
      (\x, {0.5*exp(-(\x-2.1)*(\x-2.1)/0.55)})
    -- (3.8, 0) -- (0.5, 0) -- cycle;

  \draw[teal!75!black, line width=1.5pt]
    plot[smooth, domain=1:6.5, samples=70]
      (\x, {2.6*exp(-(\x-4.9)*(\x-4.9)/0.55)});
  \node[teal!75!black, font=\footnotesize] at (5.75, 1.85) {$q_{\text{SODA}}$};

  \draw[->, blue!50!black, line width=1.2pt] (3.6, 2.9) -- (4.7, 2.9);
  \node[blue!50!black, font=\scriptsize, above=-1pt] at (4.15, 2.9) {teacher imitation};

  \draw[->, red!50!black, line width=1.2pt] (2.1, 0.55) -- (2.1, 0.08);
  \node[red!50!black, font=\scriptsize, anchor=west, align=left] at (2.4, 0.45) {mode\\pruning};

  \node[red!50!black, font=\small] at (2.1, 0.25) {\ding{55}};
\end{scope}

\end{tikzpicture}%
}
\vspace{-0.15in}
\caption{Dual learning signal in \method. \textbf{(a)}~A brief warmup shifts the student toward the teacher via imitation, but residual modes from $q_0$ persist. \textbf{(b)}~Preference-based distillation additionally suppresses these student-specific inferior behaviors via mode pruning, yielding $q_{\text{SODA}} \approx p$.}
\label{fig:dual_signal}
\end{figure}

\subsection{Algorithmic Summary}
\label{sec:algorithm}

The complete \method pipeline (\cref{alg:soda}) consists of three stages: (1)~generate base student responses $y_i^s \sim q_0(\cdot \mid x_i)$ offline to construct the preference dataset $\mathcal{D}_{\text{pref}}$; (2)~a brief warmup to obtain $q_{\text{w}}$; and (3)~preference-based distillation via DPO on $\mathcal{D}_{\text{pref}}$, starting from $q_{\text{w}}$.
\Cref{tab:compute} compares the computational profile of \method against SeqKD and GAD: \method achieves student-aware distillation without adversarial training, additional models, or continuous on-policy sampling.
We validate our design choices through ablation studies in \cref{sec:analysis}.

\subsection{Scaling Boundary and Teacher Superiority}
\label{subsec:scaling_boundary}

A core assumption of \method is the strict superiority of the teacher's responses ($y_w$) over the base student's zero-shot rollouts ($y_l$). This assumption holds strongly for our target regime of compact student models (e.g., 3B to 8B parameters), where the capability gap relative to frontier APIs remains vast. In this setting, the teacher's outputs reliably dominate, providing a clean preference signal without requiring external filtering. 

However, we explicitly acknowledge a scaling boundary: as the student model's parameter count scales up (e.g., $\ge$ 72B), its inherent capabilities begin to approximate those of the teacher. As this performance gap narrows, the student's zero-shot responses may become competitive, introducing noise into the static preference pairs. Consequently, \method is purposefully scoped and validated as an ultra-efficient distillation paradigm tailored specifically for compact models.

\section{Experiments}
\label{sec:experiments}

\subsection{Setup}
\label{sec:setup}

\noindent\textbf{Dataset and Models.}
Following \citet{ye2025black}, we evaluate on LMSYS-Chat-1M-Clean \citep{lmsys}. We adopt GPT-5-Chat \citep{gpt5} as the black-box teacher (accessed exclusively via API). For students, we use instruction-tuned models from the Qwen2.5 \citep{qwen2.5} (3B, 7B, 14B) and Llama-3 \citep{llama3} (3B, 8B) families. This exact suite enables direct comparisons across diverse scales and architectures.

\noindent\textbf{Training and Evaluation.}
\method begins with a supervised warmup on teacher responses before preference distillation. We use GPT-4o as the automated judge, adopting the evaluation protocol of \citet{ye2025black} and \citet{minillm}. Full hyperparameters and generation details are provided in Appendix \ref{appendix:implement}.

\noindent\textbf{Baselines.}
We compare against: (1)~\textbf{Base}: the original instruction-tuned student; (2)~\textbf{SeqKD} \citep{skd}: supervised fine-tuning on teacher responses, representing standard off-policy distillation and serving as our warmup phase; and (3)~\textbf{GAD} \citep{ye2025black}: the state-of-the-art fully on-policy adversarial method. For GAD, we report results directly from \citet{ye2025black} given the identical experimental setup.

\subsection{Main Results}
\label{sec:main_results}
\vspace{-0.1in}
\begin{table}[!h]
\centering
\caption{Automatic evaluation results (GPT-4o Score). Each student response is scored against a GPT-4o reference; the reported metric is $S/(S+R) \times 100$ averaged over all test prompts, where 50 indicates parity with GPT-4o. Base, SeqKD, and GAD numbers are from \citet{ye2025black}. Best result per model is in \textbf{bold}.}
\small
\begin{tabular}{ll|c|c|c|c}
\toprule
\multirow{2}{*}{Model} & \multirow{2}{*}{Method}
& \multicolumn{1}{c|}{In-Dist.} & \multicolumn{3}{c}{Out-of-Distribution} \\
& & LMSYS & Dolly & SelfInst & Vicuna \\ \midrule
GPT-5-Chat & Teacher & 51.7 & 49.8 & 49.7 & 49.9 \\ \midrule
\multirow{4}{*}{Qwen2.5-3B-Instruct}
   & Base            & 45.8 & 45.1 & 45.6 & 47.3 \\
   & SeqKD           & 47.5 & 44.8 & 45.7 & 48.0 \\
   & GAD             & 48.9 & \textbf{46.7} & 47.7 & 49.4 \\
   & \textbf{\method} & \textbf{49.2} & 46.1 & \textbf{48.2} & \textbf{49.8} \\ \midrule
\multirow{4}{*}{Qwen2.5-7B-Instruct}
   & Base            & 48.7 & 47.6 & 48.3 & 49.1 \\
   & SeqKD           & 49.2 & 47.2 & 48.3 & 49.5 \\
   & GAD             & 50.8 & 48.5 & 50.1 & \textbf{51.4} \\
   & \textbf{\method} & \textbf{51.5} & \textbf{49.6} & \textbf{50.8} & \textbf{51.4} \\ \midrule
\multirow{4}{*}{Llama-3.2-3B-Instruct}
   & Base            & 44.0 & 45.8 & 47.0 & 46.9 \\
   & SeqKD           & 47.6 & 47.0 & 47.1 & 48.1 \\
   & GAD             & 48.1 & 48.5 & 49.1 & 48.9 \\
   & \textbf{\method} & \textbf{49.1} & \textbf{49.5} & \textbf{50.5} & \textbf{49.9} \\ \midrule
\multirow{4}{*}{Llama-3.1-8B-Instruct}
   & Base            & 46.9 & 46.6 & 48.4 & 47.9 \\
   & SeqKD           & 49.7 & 47.7 & 48.7 & 48.7 \\
   & GAD             & 50.3 & 48.8 & 49.5 & 50.2 \\
   & \textbf{\method} & \textbf{51.8} & \textbf{49.9} & \textbf{51.6} & \textbf{51.9} \\
\bottomrule
\end{tabular}
\label{tab:main}
\end{table}

\Cref{tab:main} presents automatic evaluation results across five student models and four benchmarks.
Both GAD and \method substantially improve over the Base and SeqKD baselines across all settings, confirming the value of incorporating student-distribution information into black-box distillation.
The key comparison is between the two on-policy approaches: GAD and \method. While the absolute quality improvement over GAD is modest (+0.9 points on average), \method achieves strict performance parity, and often slight improvements, matching or outperforming GAD on 15 out of 16 model-dataset combinations, with up to a +2.1 margin on specific benchmarks (Llama-3.1-8B, SelfInst). These relative improvements are most visible on the Llama family, where \method maintains a lead of over 1 point across every benchmark for both 3B and 8B models. On Llama-3.1-8B, \method reaches 51.8 on LMSYS, operating within 0.1 points of the GPT-5 teacher (51.7) and slightly exceeding it on Vicuna (51.9) and SelfInst (51.6).

This competitive performance extends to out-of-distribution benchmarks, where \method consistently shows larger margins over SeqKD than GAD does. This suggests that static, preference-based error correction generalizes just as well, if not better, than continuous adversarial training on unseen prompt distributions.
Critically, \method delivers this comparable-or-better distillation quality entirely without a discriminator, adversarial training, or per-step on-policy generation (\Cref{tab:compute}). This demonstrates that a one-time snapshot of the student's distribution is sufficient to capture the alignment benefits of continuous on-policy adaptation, but at a fraction of the computational and engineering cost.


\subsection{Analysis}
\label{sec:analysis}

\noindent\textbf{Rejection source.}
The core design choice in \method is using the base student $q_0$ as the source of rejected responses.
We compare two alternatives on Qwen2.5-3B and Llama-3.2-3B, holding all other hyperparameters fixed (\Cref{tab:ablation_source}).
\textit{Cross-student} replaces $q_0$'s responses with those from a different model family's base student (Llama for Qwen and vice versa); \textit{Bad GPT-4o-mini} uses intentionally low-quality responses from GPT-4o-mini (high temperature, truncated). Both generic alternatives underperform $q_0$ by 1--2 points.
Because these sources produce generic negatives the student would never naturally generate, the optimizer learns a trivial contrast rather than penalizing the student's own innate inferior behaviors. They also require extra resources (a separate model or API calls), while $q_0$ is already available at zero extra cost.

\begin{table}[h]
\begin{minipage}[t]{0.58\textwidth}
\centering
\small

\begin{tabular*}{\linewidth}{@{\extracolsep{\fill}} l l | c | c @{}}
\toprule
Rejection source & Extra cost & Qwen-3B & Llama-3B \\
\midrule
$q_0$ (\textbf{\method}) & None & \textbf{49.2} & \textbf{49.1} \\
Cross-student & Extra model & 48.0 & 48.2 \\
Bad GPT-4o-mini & API cost & 47.7 & 46.9 \\
\bottomrule
\end{tabular*}
\vspace{-0.1in}
\caption{Rejection source ablation (GPT-4o Score, LMSYS).}
\label{tab:ablation_source}
\end{minipage}
\hfill
\begin{minipage}[t]{0.4\textwidth}
\centering
\small
\begin{tabular}{lcc}
\toprule
Method & Memory & Time \\
\midrule
GAD & 50.4 GB & $\sim$40 h \\
\textbf{\method} & \textbf{36.6 GB} & \textbf{$\sim$4 h} \\
\bottomrule
\end{tabular}
\vspace{-0.05in}
\caption{Training cost (Qwen2.5-7B, 8$\times$H100).}
\label{tab:efficiency}
\end{minipage}
\end{table}

\noindent\textbf{Efficiency.} \Cref{tab:efficiency} further shows that \method reduces per-GPU memory by 27\% and accelerates training by $\sim$10$\times$ compared to GAD, by eliminating the discriminator and per-step on-policy generation.



\begin{figure}[htb]
\centering
\includegraphics[width=0.85\linewidth]{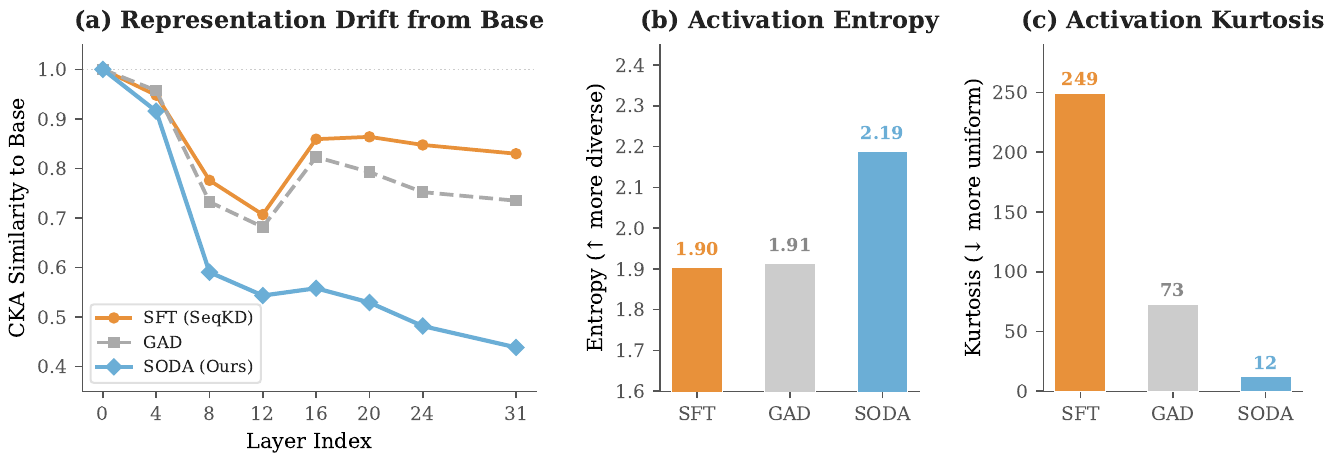}
\vspace{-0.2in}
\caption{Representation analysis on Llama-3.1-8B-Instruct (200 held-out LMSYS prompts). \textbf{(a)}~Layer-wise CKA similarity to the base model: \method diverges most, indicating deeper representational restructuring. \textbf{(b,\,c)}~Last-layer activation entropy and kurtosis: \method achieves the highest entropy and lowest kurtosis, correlating with its strongest distillation performance.}
\label{fig:repr}
\end{figure}

\noindent\textbf{Representation analysis.}
To understand how each distillation method reshapes the student's internal representations, we extract last-token hidden states from Llama-3.1-8B-Instruct (Base, SFT, \method, GAD) on 200 held-out LMSYS prompts and compute three metrics (\Cref{fig:repr}): (i) \emph{Linear CKA} \citep{kornblith2019similarity} measures representational similarity between two models at a given layer: $\text{CKA}(X, Y) = {\|X^\top Y\|_F^2}\big/\bigl({\|X^\top X\|_F \cdot \|Y^\top Y\|_F}\bigr)$, where $X, Y \!\in\! \mathbb{R}^{n \times d}$ are centered hidden-state matrices and $\|\cdot\|_F$ is the Frobenius norm; CKA $\!=\! 1$ means identical structure.
For the final hidden layer, we additionally compute two activation statistics over the flattened hidden-state values, following \citet{zhang2025rl}, who find that higher entropy and lower kurtosis correlate with stronger generalization:
(ii) \emph{activation entropy} over a histogram of all activation values (higher $=$ more diverse), and (iii) \emph{activation kurtosis} (higher $=$ a few dimensions dominate).

Three findings emerge.
First, \method drives the deepest representational restructuring, with its final-layer CKA dropping to 0.44, far below SFT and GAD (\Cref{fig:repr}a).
Second, while SFT suffers from severe representational over-specialization (kurtosis spiking to 249 vs.\ 88 for the base model), \method reduces kurtosis to 12, significantly outperforming GAD (73) (\Cref{fig:repr}c).
Third, \method uniquely raises activation entropy above the base model (2.19 vs.\ 2.08), whereas SFT and GAD both decrease it (\Cref{fig:repr}b).
These results highlight the synergy of our approach: coupling a stable preference objective with a targeted snapshot of innate inferior responses avoids the instability of adversarial training. By explicitly penalizing these natural inferior behaviors, \method induces a healthier, more diverse feature space, explaining its superior performance despite a vastly simplified pipeline.

\paragraph{Impact of Dynamic Resampling and Snapshot Staleness.} 
A natural concern regarding the semi on-policy paradigm is the ``staleness'' of the static snapshot: as the student policy $q_\theta$ diverges from the base prior $q_0$ during training, the rejected samples become increasingly out-of-distribution. Intuitively, one might assume that dynamically resampling the student's outputs to provide fresh, strictly on-policy negatives would yield better alignment gradients. 

To investigate this, we conducted an ablation study where we dynamically resampled the student policy outputs at mid-training checkpoints (Step 100 and Step 200) to construct updated preference pairs. As shown in Table~\ref{tab:resampling_ablation}, dynamic resampling consistently degrades distillation performance.

\begin{table}[h]
\centering
\small
\begin{tabular}{lccc}
\toprule
\textbf{Model} & \textbf{SODA (Static $q_0$)} & \textbf{Resample @ Step 100} & \textbf{Resample @ Step 200} \\
\midrule
Qwen2.5-3B     & \textbf{49.2}               & 48.5                         & 47.8                         \\
Llama-3.2-3B   & \textbf{49.1}               & 48.1                         & 47.4                         \\
\bottomrule
\end{tabular}
\caption{Ablation on dynamic resampling (GPT-4o Score, LMSYS). Updating the negative samples mid-training strictly degrades performance compared to retaining the initial static snapshot.}
\label{tab:resampling_ablation}
\end{table}

This consistent performance degradation empirically validates our static design choice. As the student policy improves and absorbs the teacher's distribution, the newly generated rejected samples lose their strict inferiority. Consequently, the contrastive margin between the teacher's response and the partially-aligned student's response narrows, diluting the corrective gradient. Retaining the initial static negatives mathematically anchors the penalty to the purest form of the student's innate errors. Furthermore, the increasing ``staleness'' of these samples serves as a deliberate regularization mechanism: it naturally decays the optimization gradient as the policy diverges from the prior, effectively preventing the reward hacking and over-optimization typically observed in prolonged preference training.

\section{Conclusion}
In this work, we introduced \method, a lightweight and highly efficient semi on-policy framework for black-box knowledge distillation. By leveraging a one-time static snapshot of the base student's prior as a targeted contrastive signal, \method achieves effective error correction without the prohibitive overhead of continuous adversarial training. Extensive evaluations demonstrate that \method matches or exceeds the performance of state-of-the-art fully on-policy methods while being 10$\times$ faster and consuming 27\% less memory. Ultimately, our findings reveal that the specificity of the alignment signal to the student's innate errors matters far more than continuous online sampling, offering a practical and scalable path for distilling capable small models.
\bibliography{colm2026_conference}
\bibliographystyle{colm2026_conference}

\appendix

\section{Algorithm and Method Comparison}
\label{app:algorithm}








\begin{algorithm}[h]
\caption{\method: Semi On-Policy Black-Box Distillation}
\label{alg:soda}
\begin{algorithmic}[1]
\REQUIRE Black-box teacher data $\mathcal{T} = \{(x_i, y_i^t)\}_{i=1}^{N}$, base student $q_0$, DPO temperature $\beta$
\ENSURE Distilled student model $q_{\text{SODA}}$

\STATE \textbf{\textit{Static Signal Construction \& Policy Initialization}}
\FOR{$i = 1, \ldots, N$}
    \STATE Base student rollout $\, y_i^s \sim q_0(\cdot \mid x_i)$ \hfill {\small \textcolor{gray}{// Offline batched inference}}
\ENDFOR
\STATE Construct preference dataset $\mathcal{D}_{\text{pref}} \leftarrow \{(x_i, \, y^+ \!=\! y_i^t, \, y^- \!=\! y_i^s)\}_{i=1}^N$
\STATE Warmup: $q_{\text{w}} \leftarrow \arg\min_\theta \, \mathcal{L}_{\text{SFT}}(\theta)$ starting from $q_0$ \hfill {\small \textcolor{gray}{// \cref{eq:sft}, can run in parallel}}

\vspace{0.08in}

\STATE \textbf{\textit{Semi On-Policy Preference Distillation}} \hfill {\small \textcolor{gray}{// Core alignment step}}
\STATE $q_{\text{SODA}} \leftarrow \arg\min_\theta \, \mathcal{L}_{\text{DPO}}(\theta)$ on $\mathcal{D}_{\text{pref}}$ starting from $q_{\text{w}}$, with $q_{\text{ref}} \!=\! q_{\text{w}}$ \hfill {\small \textcolor{gray}{// \cref{eq:dpo}}}

\RETURN $q_{\text{SODA}}$
\end{algorithmic}
\end{algorithm}


\section{Implementation Details}\label{appendix:implement}

\paragraph{Dataset.} Our dataset construction and evaluation protocol strictly follow \cite{ye2025black}. Specifically, we utilize their publicly available dataset\footnote{\url{https://huggingface.co/datasets/ytz20/LMSYS-Chat-GPT-5-Chat-Response}}, which comprises approximately 192K instruction prompts paired with teacher responses generated via the OpenAI API (GPT-5-Chat). Consistent with \citet{ye2025black}, we reserve 500 samples from this corpus as our primary in-distribution test set. To evaluate out-of-distribution generalization, we report additional results on the Dolly \citep{dolly}, Self-Instruct \citep{self_inst}, and Vicuna \citep{vicuna} datasets.

\paragraph{Training.} \method training starts with a warmup following~\citet{ye2025black}.
In the \textbf{warmup} phase, we fine-tune each base student on teacher responses via supervised learning for 3 epochs with learning rate $5 \!\times\! 10^{-6}$, cosine schedule, effective batch size 32, and maximum sequence length 3584 tokens.
In the \textbf{preference distillation} phase, starting from the best warmup checkpoint, we train for 1 epoch with the same learning rate schedule and $\beta= 0.1$.
The warmup model $q_{\text{SFT}}$ serves as both the initialization and the reference policy $q_{\text{ref}}$.
All training uses Fully Sharded Data Parallel (FSDP) with bfloat16 mixed precision on 8 NVIDIA H100 GPUs.
As a prerequisite, we generate one response per training prompt from each base student $q_0$ \emph{before any fine-tuning}, using vLLM {\citep{kwon2023vllm}} with temperature 0.7 and a maximum generation length of 1536 tokens.
This step is embarrassingly parallel and completes in under 30 minutes per model on 4 GPUs, adding negligible overhead compared to training.

\paragraph{Evaluation.} For each test prompt, we first generate a reference response from GPT-4o.
The student response and the GPT-4o reference are then presented pairwise to GPT-4o, which rates each on a 1--10 scale for helpfulness, relevance, accuracy, and level of detail.
We report the \textbf{GPT-4o Score}: $\frac{1}{N}\sum_{i=1}^{N} \frac{S_i}{S_i + R_i} \times 100$, where $S_i$ and $R_i$ are the student and reference scores for prompt~$i$; a score of 50 indicates parity with the GPT-4o reference.
To mitigate position bias in LLM-based evaluation {\citep{wang2024positionbias}}, we evaluate each prompt in both presentation orders (student-first and reference-first) and average the resulting scores.
All student responses are generated with greedy decoding and a maximum length of 1536 tokens. We follow the prompt templates of \citet{ye2025black}. 

\section*{LLM Usage Disclosure}
During the preparation of this work, the authors utilized LLMs to polish the manuscript's prose and provide coding assistance for implementation and data visualization. The authors have reviewed and edited all AI-generated suggestions and take full responsibility for the final content of the paper.

\Cref{fig:prompt_wrapper} shows the instruction wrapper used for both training and evaluation; \Cref{fig:prompt_eval} shows the GPT-4o scoring prompt.

\begin{figure}[h]
\begin{tcolorbox}[colback=gray!3, colframe=gray!50, boxrule=0.4pt, left=6pt, right=6pt, top=4pt, bottom=4pt]
\small\ttfamily
Below is an instruction that describes a task. \\
Write a response that appropriately completes the request. \\ \\
\#\#\# Instruction: \\
\{instruction\} \\ \\
\#\#\# Response:
\end{tcolorbox}
\caption{Prompt wrapper for training and evaluation.}
\label{fig:prompt_wrapper}
\end{figure}

\begin{figure}[h]
\begin{tcolorbox}[colback=gray!3, colframe=gray!50, boxrule=0.4pt, left=6pt, right=6pt, top=4pt, bottom=4pt]
\small\ttfamily
[User Instruction] \\
\{instruction\} \\ \\
{[Assistant 1's Response]} \\
\{response\_1\} \\ \\
{[Assistant 2's Response]} \\
\{response\_2\} \\ \\
\normalfont\small
We would like to request your feedback on the performance of two AI assistants in response to the user instruction and input displayed above.

Please rate the helpfulness, relevance, accuracy, and level of detail of their responses. Each assistant receives an overall score on a scale of 1 to 10, where a higher score indicates better overall performance.

Please first output a single line containing only two values indicating the scores for Assistant 1 and 2, respectively. The two scores are separated by a space.

In the subsequent line, please provide a comprehensive explanation of your evaluation, avoiding any potential bias and ensuring that the order in which the responses were presented does not affect your judgment.
\end{tcolorbox}
\caption{GPT-4o evaluation prompt, following \citet{ye2025black} and {\citet{minillm}}.}
\label{fig:prompt_eval}
\end{figure}

\end{document}